\documentclass[journal]{vgtc}                     

\onlineid{5515}

\vgtccategory{Research}
\vgtcpapertype{algorithm/technique}

\title{A Spectral Decomposition Framework for \texorpdfstring{\\}{ } Multiscale Nonlinear Dimensionality Reduction}

\author{%
  \authororcid{Zeyang Huang}{0000-0003-3945-1274},
  \authororcid{Angelos Chatzimparmpas}{0000-0002-9079-2376},
  \authororcid{Thomas H\"{o}llt}{0000-0001-8125-1650}, and
  \authororcid{Takanori Fujiwara}{0000-0002-6382-2752}
}

\authorfooter{
  \item
  	Zeyang Huang is with Link{\"o}ping University, Sweden. E-mail: zeyang.huang@liu.se.
  \item
	Angelos Chatzimparmpas is with Utrecht University, The Netherlands.
  	E-mail: a.chatzimparmpas@uu.nl
  \item 	Thomas Höllt is with the University of Bergen, Norway.
  	E-mail: thomas.hollt@uib.no
  \item 	Takanori Fujiwara is with the University of Arizona, United States. 
   E-mail: tfujiwara@arizona.edu.
}

\abstract{%
  We address two longstanding trade-offs in dimensionality reduction (DR).
First, preserving local neighborhoods can come at the cost of global structure. 
Neighbor embedding methods such as t-SNE and UMAP prioritize local similarity preservation but do not explicitly constrain global organization, whereas standard spectral methods such as Laplacian Eigenmaps capture smooth, coarse-scale graph structure but offer limited flexibility to depict finer local structure.
Second, the flexibility of nonlinear DR methods often comes at the cost of analytical transparency. Many methods do not explicitly reveal how high-dimensional structure produces patterns in the embedding.
We introduce \sysname{} (spectral decomposition for multiscale projection), a nonlinear DR framework built on an explicit spectral decomposition.
We express each embedding dimension as a weighted combination of spectral modes (Laplacian eigenvectors of the neighbor graph) and learn the weights using a UMAP-style cross-entropy objective.
By progressively adding spectral modes that capture finer graph structure, \sysname{} produces a sequence of embeddings, making the evolving balance between global organization and local detail explicit, controllable, and inspectable.
The explicit decomposition also reveals how each spectral mode contributes to the overall embedding and to the positions of individual points.
Quantitative evaluations on synthetic, image, and single-cell data show competitive local and global structure preservation, while the use cases illustrate how the decomposition supports interpretation of clusters and developmental trajectories at different stages of the progressive embedding process.
The project page includes an interactive demo and animations of progressive embeddings: \href{https://zeyh.github.io/SDMP/}{https://zeyh.github.io/SDMP/}.

}


\graphicspath{{figs/}{figures/}{pictures/}{images/}{./}} 

\usepackage{tabu}                      
\usepackage{booktabs}                  
\usepackage{lipsum}                    
\usepackage{mwe}                       
\usepackage{ccicons}                   
\usepackage{wrapfig}

\usepackage{algorithm}
\usepackage{algpseudocode}

\usepackage{colortbl}

\usepackage{mathptmx}                  

\usepackage{float}
\usepackage{amsmath,amsfonts,amssymb}
\newcommand{\norm}[1]{\left\lVert#1\right\rVert}

\usepackage{url}
\usepackage{comment}
\usepackage{siunitx}

\usepackage{xcolor}
\definecolor{diff}{rgb}{0,0,0} 

\newcommand{\sysname}{\textsc{SDMP}}

\usepackage{multicol}
\usepackage[font={scriptsize,sf}]{caption}
\begin{document}


\firstsection{Introduction}
\maketitle

\setlength{\abovedisplayskip}{3pt}
\setlength{\belowdisplayskip}{2pt}

\newcommand{\OriginalData}{\mathbf{X}}
\newcommand{\Layout}{\mathbf{Y}}
\newcommand{\ProjMat}{\mathbf{P}}
\newcommand{\ProjMatElem}{p}
\newcommand{\NormalizedEmbedData}{\hat{\EmbedData}}
\newcommand{\Graph}{G}
\newcommand{\nPoints}{N}
\newcommand{\nDimensions}{M}
\newcommand{\nTopEigvals}{S}
\newcommand{\nLayoutDimensions}{M'}

\newcommand{\AdjMat}{\mathbf{W}}
\newcommand{\AdjElm}[2]{w_{{#1},{#2}}}
\newcommand{\DegreeMat}{\mathbf{D}}
\newcommand{\DegreeElm}[1]{d_{#1}}
\newcommand{\LapMat}{\mathbf{L}}
\newcommand{\IdentityMat}{\mathbf{I}}
\newcommand{\Eigvecs}{\mathbf{U}}
\newcommand{\Eigvec}[1]{\mathbf{u}_{#1}}
\newcommand{\EigvecElement}[1]{u_{#1}}
\newcommand{\Eigvals}{\Lambda}
\newcommand{\Eigval}[1]{\mathbf{\lambda}_{#1}}
\newcommand{\Weights}{\mathbf{v}}
\newcommand{\Weight}{v}
\newcommand{\Regu}

\newcommand{\error}{\mathcal{E}}
\newcommand{\stage}[1]{S_{#1}} 

\newcommand{\EncoderNetwork}{f_\theta}
\newcommand{\ProjectorNetwork}{g_\theta}
\newcommand{\Centroid}{\mathbf{C}}
\newcommand{\Members}{\mathcal{P}}
\newcommand{\Loss}{\mathcal{L}}

Dimensionality reduction (DR) methods are widely used to visualize high-dimensional data, such as gene-expression data~\cite{becht2019dimensionality}, images~\cite{xia2021cluster}, or feature descriptors~\cite{jeon2025unveiling}. 
High-dimensional data may contain broad relationships among groups, manifold-level geometry, and fine neighborhood structure, which a low-dimensional embedding cannot generally preserve equally well~\cite{jeon2025unveiling,huang2022pacmapbio}.
This tension between preserving broad organization and fine neighborhood structure is referred to as the \textit{global--local} trade-off~\cite{mcinnes_umap_2020,wang2021pacmap,kury2025dreams}.

Among existing approaches, neighbor embeddings such as t-SNE~\cite{maaten2008visualizing} and UMAP~\cite{mcinnes_umap_2020} are particularly popular.  
These methods construct a weighted neighbor graph in the high-dimensional space and optimize a low-dimensional embedding to preserve local similarities.
Although graph-construction parameters affect which local relationships are encoded, the subsequent optimization adjusts point coordinates to preserve those relationships without explicitly representing global and local structure at different scales.
Consequently, separated clusters seen in t-SNE or UMAP embeddings may result partly from optimization pulling neighboring points into compact regions, instead of solely from distinct group structure in the high-dimensional data.
Recent studies also suggest that the global organization observed in t-SNE and UMAP embeddings can arise in substantial part from spectral initialization and early optimization dynamics~\cite{kobak2021initialization, huang2022pacmapbio}.

B\"{o}hm et al.~\cite{bohm2022attraction} characterize this global--local trade-off through the relative balance between attractive and repulsive forces, showing that different force balances can favor continuous manifold organization or compact cluster separation. 
Subsequent work further explains how sampling and other optimization choices shift this balance~\cite{damrich2022tsne}. 
However, the force balance does not directly indicate how structures at different scales are expressed in the resulting embedding.

A second challenge is the trade-off between \textit{optimization flexibility} and \textit{analytical transparency}.
Many nonlinear DR methods flexibly optimize point coordinates, but the resulting coordinates do not explicitly reveal how the neighborhood structure and optimization process jointly produce visible patterns.
Thus, visual analytics methods often analyze embeddings post hoc, for example through parameter sensitivity analysis or by relating visible low-dimensional structures to high-dimensional features~\cite{jeon2025unveiling,Salmanian2024DimVis,Fujiwara2020Supporting}.
Although useful for interpreting the output, these analyses are not designed to explain how the embedding process generated the observed patterns. 
Understanding how these patterns are generated matters because analysts often use them to draw conclusions about relationships in the original data.
Spectral methods such as Laplacian Eigenmaps (LE)~\cite{belkin2003laplacian} offer greater analytical transparency by construction. 
LE constructs its embedding directly from Laplacian eigenvectors derived from a neighbor graph, making the relationship between the graph and embedding mathematically explicit.
However, because its coordinates are fixed by the eigendecomposition, it offers limited flexibility to adjust the embedding to represent finer neighborhood relationships.

To address these two trade-offs through a common formulation, we introduce \sysname{} (spectral decomposition for multiscale projection), drawing on graph signal processing~\cite{shuman2013emerging,defferrard2016spectralgnn}.
Similar to LE, \sysname{} uses Laplacian eigenvectors derived from a neighbor graph, which we refer to as \textit{spectral modes}.
These modes describe patterns of variation over the graph. 
In modes associated with smaller eigenvalues, neighboring points tend to have similar values, corresponding to broader, more global structure. 
In modes associated with larger eigenvalues, the values can differ more strongly between neighbors, allowing finer, more local variation.
For a 2D embedding, LE directly uses the two spectral modes associated with the smallest nonzero eigenvalues as the embedding coordinates, restricting the embedding to these two modes related to global patterns.
In contrast, \sysname{} can use a larger set of spectral modes as a basis and learns their combination coefficients for each embedding dimension using a UMAP-style cross-entropy objective, allowing the embedding to incorporate both broader structure and finer local variation.
The selected modes define the \textit{spectral subspace} within which the embedding is optimized.

\sysname{} provides explicit control over the modes available to represent global and local structure by progressively expanding the spectral subspace during optimization. 
It begins with low-eigenvalue modes and adds higher-eigenvalue modes in stages, continuing optimization from the result of each preceding stage. 
This process produces a sequence of embeddings in which early stages emphasize broader organization and later stages allow progressively finer local variation.

\sysname{} supports analytical transparency by 
expressing the embedding as an explicit linear combination of spectral modes with learned coefficients,
allowing the contribution of each mode to be analyzed. 
Across the progressive sequence, analysts can examine when visible structures emerge, which modes contribute to them, and how those contributions vary across individual points. 
Analysts can also measure how closely an embedding optimized within a reduced spectral subspace matches the embedding obtained using all spectral modes. 
These capabilities support visual analysis of embeddings at both the dataset and point levels.

In summary, we introduce \sysname{} with the following contributions:
\begin{itemize}[itemsep=0pt,topsep=1pt,parsep=0pt]
\item a spectral parameterization combining spectral modes with UMAP-style neighborhood-preserving optimization, enabling explicit control over the modes available during optimization;
\item a progressive optimization strategy over expanding spectral subspaces, revealing how embeddings change from broad to fine structure;
\item dataset- and point-level analyses linking embedding patterns to spectral-mode contributions and guiding subspace selection; and
\item an empirical analysis of how spectral subspace size affects local and global structure preservation and how well low-frequency modes capture the final embedding.
\end{itemize}

\section{Background and Related Work}
\label{sec:background}

In this section, we position \sysname{} in the context of neighbor embeddings such as LE and UMAP, spectral graph methods, and approaches for interpreting DR results.

\subsection{Neighbor Embeddings and the Global--Local Tradeoff}
DR aims to extract low-dimensional embeddings (also referred to as projections or representations) of high-dimensional data while preserving meaningful structure~\cite{van2009dimensionality,cunningham2015linear}. 
Among various methods, neighbor embedding methods such as UMAP~\cite{mcinnes_umap_2020} and  t-SNE~\cite{maaten2008visualizing} focus on preserving relationships between nearby data points.
These methods typically construct a $k$-nearest neighbor ($k$NN) graph, define similarities between instances, and optimize an embedding that balances attractive forces between neighbors against repulsive forces among all points, a view unified by B\"{o}hm et al.~\cite{bohm2022attraction}. 

A substantial line of work shifts where the resulting compromise falls by modifying the objective, the sampling of pairs, or the optimization procedure. 
TriMap~\cite{amid2019trimap} imposes triplet constraints to better preserve large-scale structure; PaCMAP~\cite{wang2021pacmap} balances near, mid-range, and far pairs; densMAP~\cite{narayan_assessing_2021} augments the UMAP objective with local data density preservation; DREAMS~\cite{kury2025dreams} regularizes the t-SNE objective toward a precomputed global embedding, so that one weight interpolates between a locally and a globally faithful embedding; 
and in graph drawing, LGS~\cite{Miller2023lgs} exposes the trade-off as a neighborhood-size parameter controlling how many graph distances are preserved. 
These methods expose different global--local compromises by changing objective terms, pair sampling, or parameter values.
Although varying these parameters can produce multiple embeddings, each setting generally yields a separately optimized alternative and does not define a nested sequence of optimization spaces.

In contrast, a multiscale approach varies the resolution at which the data points themselves are represented.
HSNE~\cite{vanUnen2017HSNE}, HUMAP~\cite{Marcílio2025humap}, and UMATO~\cite{jeon2025umato} use representative points to establish a coarse embedding before introducing additional points.
Coarse levels embed smaller sets of representative points, while drilling down introduces additional points and finer detail.
Thus, their notion of scale is defined by sampling resolution, and not every point appears at every level.

\sysname{} differs from both approaches.
It keeps the data points, graph, and optimization objective fixed while progressively expanding the spectral subspace.
Every stage embeds all data points, but what changes is the range of graph variations available to the embedding.
Because each spectral subspace is contained in the next stage, the embedding from one stage provides the starting point for optimization in the expanded subspace.

Next, based on B\"{o}hm et al.\cite{bohm2022attraction}, we describe LE and UMAP as representative spectral and force-based approaches, respectively. 
\cref{table:notation} summarizes the notation used throughout the paper.

\subsection{Laplacian Eigenmaps (LE)}
\label{sec:laplacian_eigenmaps}

Let $\OriginalData = [\MakeLowercase{\OriginalData}_1 \cdots \MakeLowercase{\OriginalData}_\nPoints]^\top \in \mathbb{R}^{\nPoints \times \nDimensions}$ denote the input data, where $\nPoints$ and $\nDimensions$ are the numbers of data points and dimensions, respectively.
LE~\cite{belkin2003laplacian} approximates the manifold structure of $\OriginalData$ by constructing a neighbor graph. 
We focus on a $k$NN graph, which is commonly used in modern neighbor embedding methods, although the original formulation also considers an $\epsilon$-ball graph.
The $k$NN graph is then converted into a weighted adjacency matrix $\AdjMat \in \mathbb{R}^{\nPoints\times\nPoints}$, for example, using a heat kernel $\smash{\AdjElm{n}{m} = \exp(-\norm{\MakeLowercase{\OriginalData}_n - \MakeLowercase{\OriginalData}_m}^2 / t)}$ with parameter $t$.

A symmetric normalized Laplacian matrix is then computed as
\begin{equation}
    \LapMat = \IdentityMat - \DegreeMat^{-1/2} \AdjMat \DegreeMat^{-1/2}
\end{equation}
where $\DegreeMat$ is the diagonal degree matrix with entries 
$\DegreeElm{n} = \sum_m \AdjElm{n}{m}$ and $\IdentityMat$ is an identity matrix of size $\nPoints \times \nPoints$.

An embedding is obtained by eigendecomposition of $\LapMat$ by
\begin{equation}
\label{eq:lap_eig}
    \LapMat \Eigvec{i} = \Eigval{i}\Eigvec{i},
\end{equation}
where eigenpairs $\{ \Eigval{i}, \Eigvec{i} \}$ are ordered such that 
$0 = \Eigval{0} \leq \Eigval{1} \leq \cdots \leq \Eigval{\nPoints-1}$, and eigenvectors are orthonormal.
The first eigenvector $\Eigvec{0}$ is often referred to as the trivial eigenvector. 
The other non-trivial eigenvectors $\{\Eigvec{1}, \cdots, \Eigvec{\nPoints-1}\}$ are called \textit{spectral modes}.
These spectral modes can form a graph Fourier basis over the data points~\cite{shuman2013emerging} (refer to \cref{sec:spectral_graph_methods} for details).
A low-dimensional embedding is formed by selecting the spectral modes corresponding to the $\nTopEigvals$ smallest nonzero eigenvalues.
For visualization, typically $\nTopEigvals = 2$ is chosen to match 2D visualization spaces.
Following the ordering of spectral modes, such an embedding typically captures mainly the low-frequency properties of the graph and as such global structure.

Our framework leverages the spectral properties to relate embedding structure to graph frequencies, enabling analysis of how different components of the data influence the resulting embedding.

\begin{table}[t]
  \small
  \scriptsize
  \centering
  \caption{Summary of notation.\vspace{-6pt}}
  \label{table:notation}
  \makeatletter
\def\thickhline{%
  \noalign{\ifnum0=`}\fi\hrule \@height \thickarrayrulewidth \futurelet
   \reserved@a\@xthickhline}
\def\@xthickhline{\ifx\reserved@a\thickhline
               \vskip\doublerulesep
               \vskip-\thickarrayrulewidth
             \fi
      \ifnum0=`{\fi}}
\makeatother

\newlength{\thickarrayrulewidth}
\setlength{\thickarrayrulewidth}{2\arrayrulewidth}
\setlength{\tabcolsep}{5pt} 

\begin{tabular}{rl}
\thickhline{}
 $\nPoints$, $\nDimensions$, $\nLayoutDimensions$ & \# of data points, original dimensions, embedding dimensions\\
 $\OriginalData$, $\Layout$  & Original dataset, embedding result, data points indexed by $n$, $m$\\
 $\AdjMat$, $\LapMat$ & Weighted adjacency matrix, Laplacian matrix\\
 $\Eigvec{i}$ & $i$-th spectral mode (eigenvector)  derived from $\LapMat$\\
 $\nTopEigvals$, $\Eigvecs{}_{\nTopEigvals}$ & \# of spectral modes to keep, corresponding spectral subspace\\
\thickhline{}
\end{tabular}

  \vspace{-2mm}
\end{table}

\subsection{UMAP}
\label{sec:umap}
Similar to LE, UMAP constructs a weighted $k$NN graph $\AdjMat$ to model local relationships between data points. 
In UMAP, the edge weights are defined as:
\begin{equation}
    \AdjElm{n}{m} = \exp(~-\max(0,~\mathrm{dist}(\MakeLowercase{\OriginalData}_{n}, \MakeLowercase{\OriginalData}_{m})-\rho_{n}) ~~~ / ~~ \sigma_{n}~)
\end{equation} 
where $\rho_n$ and $\sigma_n$ are local connectivity parameters chosen to distribute data points uniformly over the manifold and $\mathrm{dist}(\cdot)$ is a distance function (the Euclidean distance by default).

UMAP then computes an embedding by minimizing a cross-entropy objective.
Let $\smash{\Layout = [\MakeLowercase{\Layout}_1 \cdots \MakeLowercase{\Layout}_\nPoints]^\top \in \mathbb{R}^{\nPoints \times \nLayoutDimensions}}$ denote the embedding coordinates, where typically $\nLayoutDimensions = 2$.
The embedding is initialized using a heuristic strategy, commonly based on LE.

In the embedding space, pairwise similarities are modeled as
\begin{equation}
\label{eq:umap_similarity_ld}
    q_{n,m} = (1 + a \norm{\MakeLowercase{\Layout}_n - \MakeLowercase{\Layout}_m}^{2b})^{-1}
\end{equation}
where $a$ and $b$ control the shape of the similarity function and are typically derived from the \texttt{min\_dist} hyperparameter.
The embedding is then optimized to minimize the cross-entropy between the high-dimensional similarities $\AdjElm{n}{m}$ and the low-dimensional similarities $q_{n,m}$.
In practice, UMAP symmetrizes the fuzzy graph and applies stochastic optimization with negative sampling to reduce calculation time.

Cross-entropy optimization is also used in other DR methods such as LargeVis~\cite{tang2016visualizing}, due to its flexibility and computational efficiency.
Our framework adopts this optimization strategy but applies it to a spectral subspace.
Specifically, we optimize an $\nLayoutDimensions$-dimensional embedding of a higher-dimensional spectral subspace (e.g., $\nTopEigvals = 20$), enabling a balance between expressiveness and explainability of the resulting embedding.

\subsection{Explanation and Analysis of DR Results}
Low-dimensional embeddings serve not only as visual summaries, but also as tools for reasoning about clusters, trajectories, and relationships among groups. 
This broader role makes their interpretation a central challenge in DR~\cite{dagstuhlembeddings2025}. 
For linear projections, the low-dimensional embedding often can be mathematically guaranteed to have a clear specific interpretation. 
For instance, in PCA, each coordinate is an explicit weighted sum of input features, and the weights are themselves the explanation. 
Nonlinear neighbor embeddings, such as t-SNE~\cite{maaten2008visualizing} and UMAP~\cite{mcinnes_umap_2020}, place points by stochastic optimization of a non-convex objective over the coordinates directly.
As a result, the final embedding exposes no parameters that account for its structure. 
Explanation can only be inferred after the optimization process.

The majority of studies focus on visual analytics approaches that support interactive inspection of uncertainty~\cite{becht2019dimensionality}, exploration of cluster relationships~\cite{xia2021cluster}, and analysis of embedding artifacts~\cite{9064929}. 
Other methods interpret regions and patterns in the embedding in terms of original high-dimensional attributes and local feature contributions~\cite{Faust2019DimReader,Coimbra2016Explaining,TIAN202193}, and characterize visual structures through logical predicates or sparse explanatory partitions~\cite{Montambault2025DimBridge}. 
Interactive systems further support steering, comparison, and task-driven analysis, enabling analysts to refine embeddings, compare cluster structure, and relate patterns to downstream analytical goals~\cite{Fujiwara2022Interactive,Liu2023RankAxis,Manz2025A,Okami2026Visual}.
These approaches support assessment and interpretation of embeddings, but remain post hoc. 

There is a smaller body of theoretical and methodological work examining the internal optimization mechanisms of nonlinear DR methods~\cite{bibal2021biot,dagstuhlembeddings2025, bohm2022attraction, cai2022theoretical}.
For example, Cai et al.~\cite{cai2022theoretical} show that the early stage of t-SNE is equivalent to power iterations on the graph Laplacian, implicitly performing spectral clustering, while later stages exhibit amplification and stabilization dynamics. 
More broadly, recent work shows that contrastive and non-contrastive representation learning objectives can often be interpreted through spectral embedding formulations~\cite{balestriero2022contrastive}.
These analyses give theoretical support for viewing the graph spectrum as the natural basis in which neighbor embeddings operate.

Another line of work proposes explanation-by-design~\cite{arrieta2020xai} DR methods~\cite{xia2021cluster}.
LPP~\cite{he2003lpp}, NPE~\cite{he2005npe}, and linear t-SNE~\cite{bunte2012tsnle, xia2021cluster} are the linear variants of LE~\cite{belkin2003laplacian}, LLE~\cite{roweis2000lle}, and
t-SNE~\cite{maaten2008visualizing}, respectively.
They construct the embedding with a single linear map of the input features and optimize that map to preserve the same local-neighborhood relationships. 
More recently, Couplet et al.~\cite{couplet2023natively} propose a natively interpretable variant of t-SNE that learns different linear feature weights for each point, while encouraging neighbor points to have similar weights, and FeatureMAP~\cite{yang2026featuremap} approximates local tangent spaces to recover feature directions and importances within a UMAP-style embedding.

These explanation-by-design methods~\cite{couplet2023natively, yang2026featuremap, bunte2012tsnle} are conceptually closest to ours because they derive explanations directly from the quantities used to construct the embedding.
Their explanations identify which input features contribute to a point's position.
In \sysname{}, each point's position is expressed in terms of spectral modes ordered from coarse to fine variation on the neighbor graph.
The contribution of each mode thus indicates the scale of graph structure it represents in the embedding.

\subsection{Spectral Graph Methods for Embedding Analysis}
\label{sec:spectral_graph_methods}

\begin{figure}[!t]
    \centering
    \includegraphics[width=1\linewidth]{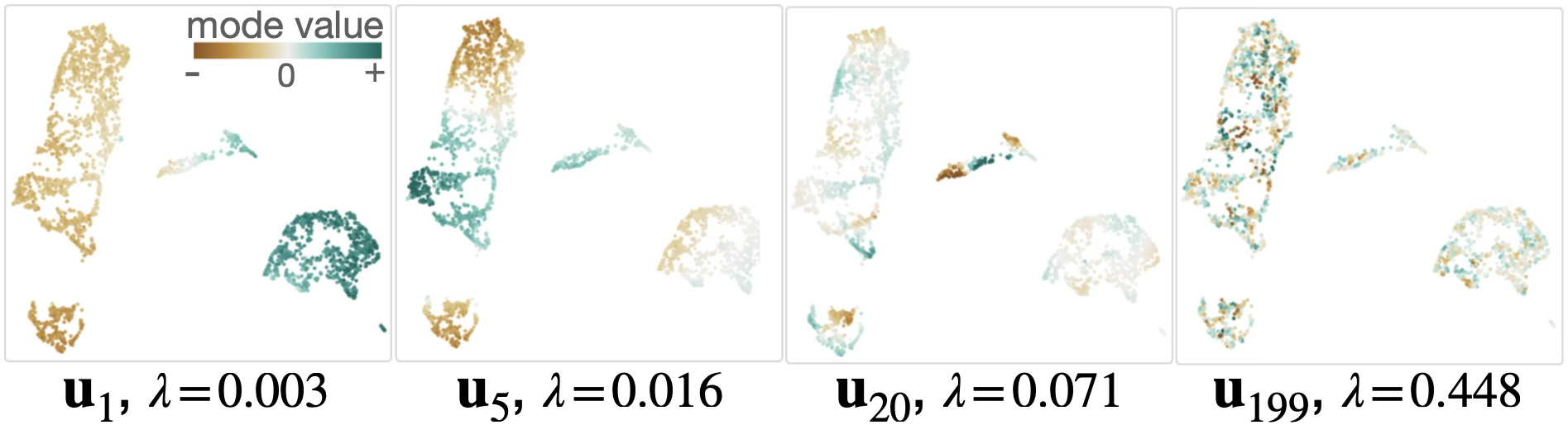}
    \caption{Example of selected spectral modes on a UMAP embedding~\cite{mcinnes_umap_2020} of sampled Fashion-MNIST data~\cite{xiao2017fashion}. 
    Points are colored by values in the corresponding mode $\Eigvec{i}$, which is an $\nPoints$-length vector.
    Low-index modes vary smoothly across the graph and form broad coherent regions (left), whereas higher-index modes carry local variation (right)~\cite{hammond2011wavelets}.}
    \label{fig:mode_explain}
\end{figure}

Laplacian eigenvectors serve directly as embedding coordinates in LE~\cite{belkin2003laplacian, hammond2011wavelets}.
More generally, spectral graph theory~\cite{spielman2012spectral} interprets these eigenvectors as an ordered basis for graph signals.
It has been widely studied in areas such as spectral graph neural networks~\cite{dong2026graph, wang2022powerfulgnn, defferrard2016spectralgnn}, spectral diffusion metrics~\cite{liao2024assessing}, graph diagnostics~\cite{tsitsulin2018netlsd, thibeault2024low}, and scientific visualization~\cite{rosen2023homology,benson2011spectral}. 
As illustrated in \cref{fig:mode_explain}, the ordering of eigenvalues provides a multiscale view of graph structure. Low-order modes move strongly connected vertices coherently, whereas higher-order modes permit increasingly rapid variation across graph edges. Spectral analysis can act as an ordered basis for measuring how an existing embedding varies relative to the graph intended to represent.

Spectral graph methods have been used for DR through diffusion-based methods that exploit the spectrum of graph operators to capture manifold structure and trajectories~\cite{coifman2006diffusion,moon_PHATE_2019,huguet2023heat}.
The connection between spectral modes and diffusion follows directly from the heat operator $H_t=e^{-t\LapMat}$, which keeps each mode $\Eigvec{i}$ but scales its magnitude by $e^{-t\lambda_i}$, such that $H_t\Eigvec{i}=e^{-t\lambda_i}\Eigvec{i}$. 
Because higher-order modes have larger eigenvalues, they are attenuated more rapidly as the diffusion time $t$ increases.
Diffusion Map~\cite{coifman2006diffusion} uses these time-weighted modes directly as embedding coordinates.  PHATE~\cite{moon_PHATE_2019} applies the analogous discrete diffusion for $t$ steps, transforms the resulting transition probabilities into diffusion potentials, and embeds their pairwise distances using multidimensional scaling (MDS).  HeatGeo~\cite{huguet2023heat} derives a heat-geodesic dissimilarity from $H_t$ and similarly embeds the resulting distances. 
MS-IMAP~\cite{deutsch2026msimap} uses spectral graph wavelets at multiple scales to learn an embedding with one coordinate per input feature, enabling feature importance estimation. 
\sysname{} shows how spectral modes contribute to each point’s position in the 2D embedding.

Diffusion methods use diffusion time $t$ to control how far information propagates along the graph.
In PHATE and HeatGeo, the final MDS axes do not correspond one-to-one with the modes used to construct the distances. 
\sysname{} parameterizes each embedding axis directly in a spectral subspace and optimizes the coefficient of each included mode under the embedding objective.
This design allows progressively adding modes that extend the available scales, while keeping their contributions directly interpretable.

Spectral graph representations have also been used in interactive and analytical systems. 
SpectralNET~\cite{Forman2005SpectralNET} extends LE into an interactive system that visualizes spectral quantities such as eigenvalues and eigenvectors. 
SpectralMAP~\cite{watanabe2023spectralmap} uses spectral decomposition of the data to construct a global filter that guides manifold approximation. 
While these approaches use spectral representations for embedding or analysis, they typically produce a single embedding without exposing how individual spectral modes influence the result. 
In contrast, our approach treats the spectral basis as a controllable subspace for optimization.

\section{Demonstrating Examples}
\label{sec:demo_examples}
\begin{figure*}[t]
    \centering
    \includegraphics[width=1\linewidth]{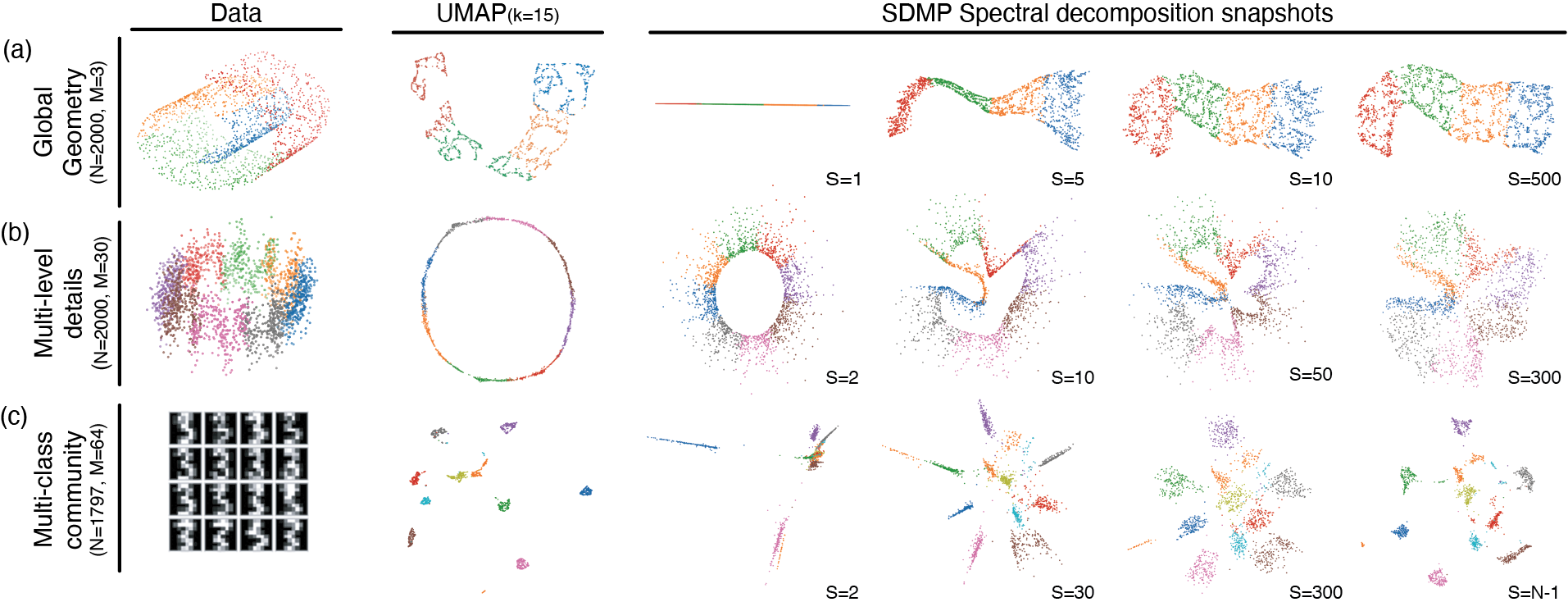}
    \caption{Comparison of data structure, UMAP~\cite{mcinnes_umap_2020}, and \sysname~results on three datasets. As the number of used spectral modes $\nTopEigvals$ increases, low-frequency snapshots produced by \sysname{} recover coarse organization first and then add finer detail toward the full-spectrum embedding.}
    \label{fig:motivating_example}
\end{figure*}

Our work approaches neighbor embeddings through the lens of spectral graph analysis. 
As discussed, neighbor embeddings consist of two key components: (1) constructing a graph representation of high-dimensional data and (2) optimizing an embedding to preserve this graph structure.
We observe that these two components serve distinct roles in data analysis. 
Spectral analysis of the graph reveals intrinsic structures across multiple scales, ranging from coarse global organization to fine local variations. 
In contrast, cross-entropy optimization arranges these structures in a low-dimensional space while preserving neighborhood relationships. 
Before we explain \sysname{} in depth in \cref{sec:method}, here we illustrate this perspective using three representative examples shown in \cref{fig:motivating_example}.

\vspace{1pt}\noindent\textbf{Datasets.}
Our examples consist of three datasets (\cref{fig:motivating_example}) chosen for their different structural characteristics:
\setlist[description]{font=\normalfont}
\begin{description}[topsep=0pt,itemsep=0pt,parsep=0pt]
    \item[1.] The Swiss Roll~\cite{tenenbaum2000global} dataset has a smooth manifold structure embedded in a higher-dimensional space. It is constructed by sampling a 2D sheet rolled up in 3D space ($\nPoints {=} 2{,}000$, $\nDimensions {=} 3$).
    \item[2.] The Multiscale Periodic Loop represents multiscale structure with a low-frequency global circular pattern and superimposed high-frequency radial oscillations and noise ($\nPoints {=} 2{,}000$, $\nDimensions {=} 30$).
    \item[3.] MNIST~\cite{deng2012mnist} represents data with a clear cluster structure. We use a version formed by $8{\times}8$ grayscale digit images, flattened into vectors ($\nPoints {=} 1{,}797$, $\nDimensions {=} 64$).
\end{description}

\vspace{1pt}\noindent\textbf{Results.}
\cref{fig:motivating_example} shows 2D embeddings (``spectral decomposition snapshots'') generated by \sysname{} and UMAP for comparison.
For each dataset, we vary the number of spectral modes $\nTopEigvals$ used, corresponding to the smallest nonzero eigenvalues (\cref{eq:lap_eig}).  
Cross-entropy optimization is then applied within the corresponding subspace.

For the Swiss Roll (\cref{fig:motivating_example}(a)), using small $\nTopEigvals$ produces an embedding that preserves the global structure as an unfolded trajectory. 
As $\nTopEigvals$ increases, the embedding begins to reflect finer local variations and noise. 
Compared to UMAP, our method better preserves manifold continuity by explicitly incorporating low-frequency spectral modes.

For the Multiscale Periodic Loop (\cref{fig:motivating_example}(b)), the embedding first captures the global circular structure. 
As additional spectral modes are introduced, finer oscillatory patterns along the loop become visible, revealing multiscale structure. 
In contrast, UMAP produces a single circular structure and does not expose these finer variations.

For the MNIST (\cref{fig:motivating_example}(c)), the results illustrate how spectral modes relate to class separability. 
Classes that form coherent neighborhoods in the graph can be separated using spectral modes with small eigenvalues, while classes with more ambiguous boundaries require spectral modes with large eigenvalues for clear separation. 
For example, some digit classes become distinguishable with only a few spectral modes (e.g., blue points), whereas others require additional spectral modes to form distinct clusters (e.g., red points). 

These examples demonstrate that \sysname{} enables explicit preservation of global structure, progressive incorporation of multiscale patterns, and analysis of community structure seen in the embedding.

\vspace{-2pt}

\section{\sysname: \textbf{S}pectral \textbf{D}ecomposition for Multisc\textbf{a}le \textbf{P}rojection of High-dimensional Data}
\label{sec:method}

We introduce \sysname{}, which provides a multiscale view of high-dimensional data.
\sysname{} separates the roles of structure representation and embedding optimization by combining a spectral basis derived from the graph representation with a cross-entropy-based embedding.

\vspace{-1pt}
\subsection{Overview}
The embedding process involves four main steps: (1) weighted $k$NN graph construction, (2) spectral modes extraction, (3) selection of a spectral subspace, and (4) cross-entropy-based optimization of a low-dimensional embedding. 
Beyond embedding, the framework also supports analysis through (5) progressive subspace construction and embedding as well as (6) visual explanation of the spectral profiles.

The first two steps follow existing neighbor embedding approaches. 
We construct a weighted $k$NN graph $\AdjMat$ as in UMAP~\cite{mcinnes_umap_2020} (i.e., a fuzzy graph), as described in \cref{sec:umap}.
We then compute spectral modes over $\AdjMat$ as described in \cref{sec:laplacian_eigenmaps}.

\sysname{} mainly differs in the subsequent steps. 
Moving away from directly optimizing the positions of data points to adjust pairwise relationships between them, we operate in an $\nTopEigvals$-dimensional spectral subspace and learn a projection mapping onto an $\nLayoutDimensions$-dimensional embedding space (typically $\nLayoutDimensions = 2$ for visualization). 
In this formulation, cross-entropy optimization is used to identify an informative embedding subspace from the spectral subspace. 
This separation enables structured control over the embedding process and facilitates analysis of how spectral modes influence the embedding.

As demonstrated in \cref{sec:demo_examples}, progressive subspace construction and embedding allow spectral modes to be added incrementally, enabling coarse-to-fine refinement of the embedding. 
Because \sysname{} expresses each mode's influence as an explicit quantity, it opens up the embedding to visual inspection at both the dataset and the data point levels.

\subsection{Spectral Subspace}

The Laplacian spectrum obtained from \cref{eq:lap_eig} provides an ordering of graph signals from coarse to fine scales~\cite{shuman2013emerging}. 
Spectral modes associated with small eigenvalues capture smooth variations over the neighbor graph, while those with larger eigenvalues encode increasingly localized structure. 
This ordering follows from the Courant--Fischer theorem~\cite{spielman2012spectral}.
It shows that the eigendecomposition of the graph Laplacian can be understood as repeatedly selecting the smoothest graph signal not already represented by the preceding modes. 
Orthogonality of eigenvectors prevents each new mode from repeating an earlier pattern, so the smallest remaining variation cannot decrease as the mode index increases. 
For a normalized eigenvector, its eigenvalue is exactly this weighted variation across graph edges, $\Eigval{i} = \Eigvec{i}^\top \LapMat \Eigvec{i}$, which accumulates a contribution from every edge in proportion to its weight.
Thus, the modes are ordered from coarse to fine by construction.
This property enables a multiscale representation of the data.
By selecting spectral modes for embedding optimization, we explicitly control the spectral bandwidth of the representation, thereby making the global--local structure trade-off explicit and adjustable.

Let $\Eigvecs{}_{\nTopEigvals} = [\Eigvec{1} \,\cdots\, \Eigvec{\nTopEigvals}] \in \mathbb{R}^{\nPoints \times \nTopEigvals}$ denote the subspace spanned by the $\nTopEigvals$ spectral modes corresponding to the smallest eigenvalues. 
When $\nTopEigvals$ is small, the subspace is restricted to low-frequency modes, and the resulting embedding captures only coarse global structure, such as connected components or manifold-level organization. 
As $\nTopEigvals$ increases, higher-frequency modes are incorporated, allowing finer local variation to appear in the embedding.
In the limit where $\nTopEigvals = \nPoints-1$, the subspace spans the full non-trivial spectrum, and the resulting embedding becomes close to those produced by conventional neighbor embedding such as UMAP.

\subsection{Embedding Space Optimization}
\label{sec:layout_optimization}

Given a selected spectral subspace $\Eigvecs{}_{\nTopEigvals}$, we learn an embedding by optimizing a projection of this subspace. 
Our method parameterizes the embedding as:
\begin{equation}
\label{eq:core_equation}
    \Layout = \Eigvecs{}_{\nTopEigvals} \cdot \ProjMat{}_{\nTopEigvals},
\end{equation}
where $\ProjMat{}_{\nTopEigvals} \in \mathbb{R}^{\nTopEigvals \times \nLayoutDimensions}$ is a projection matrix mapping the spectral subspace to an $\nLayoutDimensions$-dimensional embedding space. 

The projection factorizes into a fixed basis representation, $\OriginalData_{\MakeLowercase{\nPoints}} \rightarrow (\EigvecElement{\MakeLowercase{\nPoints},1}, \ldots, \EigvecElement{\MakeLowercase{\nPoints},\nTopEigvals})$, given by graph construction and eigendecomposition. 
The mapping from the input data to the spectral features is nonlinear because it depends on the neighborhood graph constructed from the data. 
We compute the spectral basis $\Eigvecs{}_{\nTopEigvals}$ from the graph Laplacian and keep it fixed during embedding optimization. 
The learned matrix $\ProjMat_{\nTopEigvals}$ then maps these spectral features linearly to the embedding coordinates.

To obtain $\ProjMat_{\nTopEigvals}$, we adopt the same low-dimensional similarity definition $q_{n,m}$ as UMAP (see \cref{eq:umap_similarity_ld}) and optimize the projection matrix by minimizing the cross-entropy objective:
\begin{equation}
    \mathcal{L}_s(\ProjMat{}_{\nTopEigvals})
    = -\sum_{n,m}\AdjElm{n}{m}\log q_{n,m}
    -\gamma\sum_{n,l}\log(1-q_{n,l}), 
\end{equation} 
where $\AdjElm{n}{m}$ denotes the edge weight in the weighted $k$NN graph, $l$ denotes a point not adjacent to $n$ drawn by negative sampling, and $\gamma$ controls the strength of repulsion, taken from the UMAP implementation. 
This formulation is fully differentiable once $\Eigvecs{}_{\nTopEigvals}$ is computed~\cite{damrich_umap_2021} (refer to the supplementary material for details), and we follow UMAP's negative sampling strategy for efficient optimization.
Regarding initialization, the difference is that \sysname{}'s optimization starts from $\ProjMat_{\nTopEigvals} = [\,\IdentityMat_{\nLayoutDimensions} \,;\, \mathbf{0}\,]$ perturbed by small random values, so the initial embedding is approximately the first $\nLayoutDimensions$ non-trivial spectral modes.

This reparameterization shifts the optimization from point coordinates to a low-dimensional subspace, enabling the embedding to remain expressive while providing intrinsic analytical transparency. 
In particular, the learned projection matrix $\ProjMat{}_{\nTopEigvals}$ encodes how each spectral mode contributes to the embedding.
Its $i$-th row, $\mathbf{\ProjMatElem}_i$, contains the coefficients of mode $i$ along the embedding axes.
Since the selected eigenvectors are orthonormal ($\smash{\Eigvecs_{\nTopEigvals}^\top\Eigvecs_{\nTopEigvals} = \IdentityMat_{\nTopEigvals}}$), Parseval's identity~\cite{shuman2013emerging} ensures that this decomposition does not double-count the embedding and gives
$\|\Layout\|_F^2 = \|\ProjMat_{\nTopEigvals}\|_F^2
= \sum_{i=1}^{\nTopEigvals}\|\mathbf{\ProjMatElem}_i\|_2^2$
, where $\|\cdot\|_F$ denotes the Frobenius norm.
Thus, $\|\mathbf{\ProjMatElem}_i\|_2^2/\|\Layout\|_F^2$ is the fraction of the embedding's total squared norm contributed by mode $i$.
We visualize these fractions as a spectral response profile to compare how embeddings distribute their contributions across graph scales (\cref{sec:visual_explaination}).

In summary, \sysname{} reformulates embedding optimization as learning a projection from a spectral subspace, combining the flexibility of optimization-based methods with the structured and analyzable representation provided by spectral analysis.

\subsection{Progressive Embedding Strategy}
\label{sec:progressive_strategy}

The spectral reparameterization of the cross-entropy objective enables optimization within a progressively expanding spectral subspace:
\begin{equation}
    \Eigvecs{}_{s_1}\subset \Eigvecs{}_{s_2}\subset\cdots\subset  \Eigvecs{}_{s_T} = \Eigvecs{}_{\nTopEigvals},\quad
    s_1<s_2<\cdots<s_T
\end{equation}
where the schedule $\{s_r\}_{r=1}^T$ defines how the dimensionality of the spectral subspace is increased, with $s_r$ denoting the number of spectral modes included at stage $r$.

At the initial stage ($r = 1$), we optimize the embedding using $\Eigvecs{}_{s_1}$ and obtain a projection matrix $\ProjMat{}_{s_1}$. 
At each subsequent stage ($r = 2, \ldots, T$), we expand the subspace to $\Eigvecs{}_{s_r}$ and initialize the projection matrix by augmenting the previous solution, 
    $\ProjMat{}_{s_r} = [\, \ProjMat{}_{s_{r-1}} \,;\, \delta_{s_r} \,]$
where $\delta_{s_r} \in \mathbb{R}^{(s_r - s_{r-1}) \times \nLayoutDimensions}$ contains small random values. 
Optimization then proceeds within the expanded subspace. 
This process is repeated until the intended subspace $\Eigvecs{}_{\nTopEigvals}$ is reached.
While more sophisticated stage-allocation strategies could be incorporated, in practice, given a total number of optimization epochs \texttt{n\_epochs}, we allocate $\lfloor \texttt{n\_epochs} / T \rfloor$ epochs to each stage.
The progressive schedule also motivates changing the attraction--repulsion update rule. Early stages restrict the embedding to few modes, so graph neighbors can be placed far apart in the embedding space, where an attraction that decays with distance gives little force to pull them back in later stages. Thus, we multiply the attractive update of a positive pair by the current distance $\norm{\MakeLowercase{\Layout}_n - \MakeLowercase{\Layout}_m}$ and increase the repulsion strength, which also makes the \sysname{} non-progressive full-spectrum run differ from UMAP.

This strategy incorporates spectral modes from coarse to fine scales during optimization. 
Early stages, constrained to low-frequency modes, capture global organization, while later stages progressively introduce higher-frequency modes that refine local structure.
As a result, the optimization produces not only a final embedding $\Layout_{s_T}$ but also a sequence of intermediate embeddings $[\Layout_{s_1}, \ldots, \Layout_{s_T}]$.

\vspace{5pt}
\noindent\textbf{Approximate spectral updates.}
To apply this strategy to larger datasets without computing or storing a full spectral basis, \sysname{} can optionally approximate spectral updates. 
While the active subspace is small, we compute its low-frequency modes explicitly and project each update onto that subspace exactly. 
As the target mode count grows, we use approximate low-pass filtering to admit finer-scale changes without computing additional modes. 
At the full-spectrum endpoint, we recover an exact projection by removing only the update component along the excluded first eigenvector.

For the filtered stages, we use stochastic Lanczos quadrature to estimate the number of modes below a given eigenvalue and select a cutoff corresponding to approximately $s_r$ modes~\cite{chen2021analysis}.
We preserve the part of the update along the low-frequency modes computed earlier and filter the rest. 
We approximate the filter with a Chebyshev polynomial and apply it through repeated multiplication by the sparse Laplacian~\cite{hammond2011wavelets}.
We evaluate the computational benefits of this approximation in \cref{sec:complexity_analysis}.

\subsection{Visual Explanations of the Embedding} 
\label{sec:visual_explaination}

As shown in \cref{eq:core_equation}, \sysname{} represents the embedding in a spectral basis, enabling intrinsic explanations at the dataset and point levels.
The \href{https://zeyh.github.io/SDMP/}{project page} provides a prototype visual analytics system that integrates progressive embedding playback with dataset- and point-level spectral explanations. 
Readers can examine how modes contribute to the overall embedding at each stage and select a point to inspect how those modes contribute to its position.

\vspace{3pt}
\noindent\textbf{Global dataset-level explanations (\textbf{G1, G2}).} 
Each column of $\Eigvecs{}_{\nTopEigvals}$ represents a spectral mode, with the columns ordered from low to high frequency.
The corresponding rows of $\ProjMat{}_{\nTopEigvals}$ contain that mode's learned coefficients for the embedding dimensions.
We derive two dataset-level explanations:
\setlist[description]{font=\normalfont}
\begin{description}[topsep=0pt,itemsep=0pt,parsep=0pt]
    \item [\textbf{G1}.] \textit{reconstruction error} $\error_{\nTopEigvals}$, measuring how closely the first $\nTopEigvals$ modes reconstruct the full-spectrum embedding;
    \item[\textbf{G2}.] \textit{response magnitude} $\|\mathbf{\ProjMatElem}_i\|_2^2$ for mode $i$, measuring how strongly that mode is weighted in the embedding.
\end{description}

For \textbf{G1}, we define the reconstruction error as 
\begin{equation}
\label{eq:reconstruction_error}
    \error_{\nTopEigvals}
    :=
    \|\Layout_{\mathrm{final}}
    -\widehat{\Layout}_{\nTopEigvals}\|_{F},
\end{equation}
where $\Layout_{\mathrm{final}}$ is obtained using all nontrivial spectral modes and $\widehat{\Layout}_{\nTopEigvals}$ reconstructs it using only the first $\nTopEigvals$ modes and their corresponding final coefficients.
Plotting $\error_{\nTopEigvals}$ as $\nTopEigvals$ increases produces a reconstruction curve for each dataset.
The curve helps analysts identify the smallest spectral subspace that approximates the final embedding within a chosen tolerance.
Examining the curve together with the corresponding reconstructed embeddings also shows when prominent patterns in the final embedding begin to appear and which patterns require additional, higher-frequency modes.
We demonstrate the evaluation and analysis use of \textbf{G1} in \cref{sec:reconstruction_error,sec:usecase_mnist}.

For \textbf{G2}, the response magnitudes across modes are shown as a \textit{response profile}, revealing how the embedding's learned weight is distributed over the spectrum.
As demonstrated in \cref{sec:usecase_mnist}, comparing these profiles across datasets reveals whether their embeddings rely primarily on smoother modes or draw on a broader portion of the spectrum.

\vspace{3pt}
\noindent\textbf{Local point-level explanations (\textbf{L1, L2}).}
Each row of $\Eigvecs{}_{\nTopEigvals}$ corresponds to a data point, and each entry $\EigvecElement{n,i}$ represents the signed participation of point $n$ in mode $i$. 
Together with the corresponding coefficient vector $\mathbf{\ProjMatElem}_i$, this entry determines the mode's contribution $\EigvecElement{n,i}\mathbf{\ProjMatElem}_i$ to the point's embedding coordinates.
For point $n$ and each included mode $i=1,\ldots,\nTopEigvals$, we derive two point-level explanations:
\setlist[description]{font=\normalfont}
\begin{description}[topsep=0pt,itemsep=0pt,parsep=0pt]
    \item[\textbf{L1}.] \textit{participation magnitude} $|\EigvecElement{n,i}|$, indicating how strongly point $n$ participates in the spectral mode independently of embedding optimization;
    \item[\textbf{L2}.] \textit{displacement magnitude} $\norm{\EigvecElement{n,i}\mathbf{\ProjMatElem}_i}_2$, indicating how strongly that mode contributes to point $n$'s final embedding coordinates after applying its learned coefficient.
\end{description}

\textbf{L1} and \textbf{L2} share the same points-by-modes structure and can be displayed as paired heatmaps with rows grouped by class or region
(\cref{fig:expl_point}). 
For each point and mode, \textbf{L2} equals \textbf{L1} scaled by the learned coefficient magnitude:
$\norm{\EigvecElement{n,i}\mathbf{\ProjMatElem}_i}_2
=|\EigvecElement{n,i}|\norm{\mathbf{\ProjMatElem}_i}_2$.
Thus, the \textbf{L1} heatmap shows which points participate strongly in each graph mode, while the \textbf{L2} heatmap shows how strongly those point-level patterns contribute to the final embedding after applying the learned mode weights.
Comparing them reveals which graph-derived patterns are emphasized or attenuated in the embedding.
When rows are grouped by class or region, the same heatmaps also reveal which modes contribute most strongly to particular groups.
We demonstrate these point- and group-level explanations in \cref{sec:usecase_mnist}.

Embeddings obtained with approximate spectral updates also support explanations of selected modes.
We reuse cached eigenvectors and compute additional modes on demand from the same graph Laplacian.
For uncached modes, we can, for example, use SciPy's shift-invert eigensolver to target eigenvectors near the estimated eigenvalue of the mode of interest.
For each embedding axis, we obtain a mode's coefficient by summing the products of the saved embedding coordinates and the corresponding entries of its unit-length eigenvector. 
This enables explanations of selected modes without constructing the full spectral basis or rerunning embedding optimization.
Integrating this on-demand computation into an interactive visual analytics system is a direction for future work.

\section{Evaluation}
\label{sec:results}

We evaluate \sysname{} from three perspectives: spectral reconstruction and selection of the subspace size (\cref{sec:reconstruction_error}), embedding quality across progressive stages in comparison with baseline DR methods (\cref{sec:quantitative}), and runtime and scalability (\cref{sec:complexity_analysis}).

\cref{tab:datasets} summarizes the datasets used for controlled comparisons, spanning synthetic data with multiscale structure,  image benchmarks (MNIST and Fashion-MNIST), and a single-cell RNA-sequencing dataset of \textit{C.~elegans} embryogenesis~\cite{packer2019lineage}.
For \textit{C.~elegans}, we evaluate both the original 20,220-dimensional data and a 100-dimensional PCA representation obtained through standard preprocessing following Monocle~\cite{trapnell2014dynamics}.
For controlled comparisons, we randomly subsample each dataset to 5,000 points to keep runtime and memory requirements manageable across methods, including MDS~\cite{torgerson1952multidimensional}, LE~\cite{belkin2003laplacian}, PHATE~\cite{moon_PHATE_2019}, 
UMAP~\cite{mcinnes_umap_2020},
TriMap~\cite{amid2019trimap},  PaCMAP~\cite{wang2021pacmap}, and the non-approximate \sysname{}.
The same dataset size also helps compare progressive embeddings at the same spectral mode counts.

Unless otherwise stated, SDMP uses a fuzzy 15-NN graph and a total of 500 optimization epochs.
In progressive runs, the number of included nontrivial spectral modes increases over ten approximately logarithmically spaced stages (i.e., $\nTopEigvals = 2, 5, 11, \ldots, 4999$).
For the embedding quality comparison (\cref{sec:quantitative}), we also run a non-progressive full-spectrum optimization, denoted by $\stage{\mathrm{full}}$, with the same epoch budget.
We use the default parameter settings for the baseline methods.

The project page shows animations of progressive embeddings, with point positions linearly interpolated between stages. Examples include the full MNIST and Fashion-MNIST datasets (70,000 images each) and a mouse-brain dataset (1.3 million cells).

\subsection{Reconstruction Error Analysis}
\label{sec:reconstruction_error}
\sysname{} constructs an embedding progressively within spectral subspaces of increasing size $\nTopEigvals$. 
A practical question is how many low-frequency modes are needed to reconstruct the final full-spectrum embedding within a chosen tolerance.
We use the first $\nTopEigvals$ modes with their coefficients from the full-spectrum embedding to compute the reconstruction error in~\cref{eq:reconstruction_error}.
We normalize the reconstruction error by the single-mode error, $\error_{\nTopEigvals}/\error_{1}$, so that the normalized error equals $1$ when $\nTopEigvals=1$ and reaches $0$ when all nontrivial spectrum modes are used.
This normalized error serves two analytical purposes.
First, its rate of decrease characterizes the spectral compressibility of the embedding. A rapid decrease indicates that the full-spectrum embedding can be substantially approximated using few modes.
Second, it identifies the smallest $\nTopEigvals$ that satisfies a chosen reconstruction-error tolerance, thereby guiding spectral subspace selection.
Thus, $\error_{\nTopEigvals}$ quantifies the discrepancy introduced by restricting the embedding to the first $\nTopEigvals$ modes.

\begin{figure}[t]
\centering
    \includegraphics[width=1\linewidth]{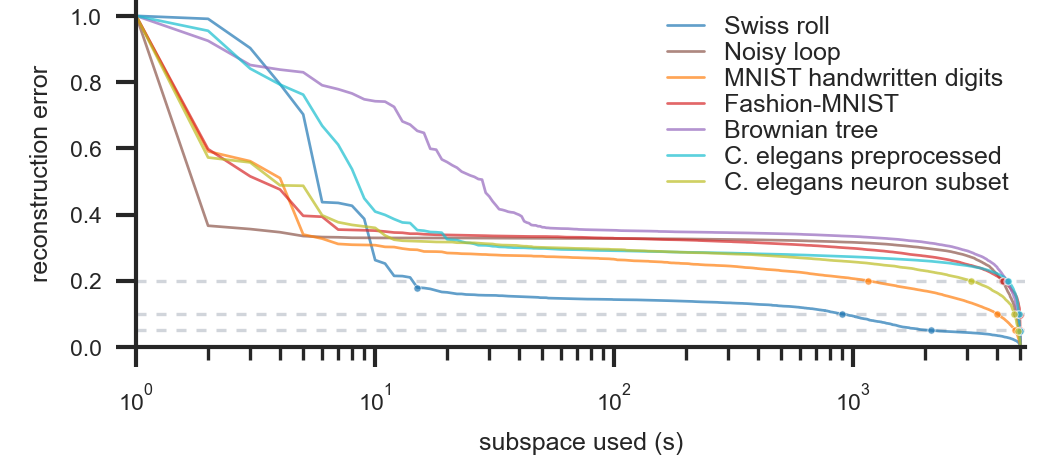}
    \caption{
    Normalized reconstruction error across datasets as the spectral subspace size $\nTopEigvals$ increases.
    The curves measure reconstruction error relative to the full-spectrum embedding.
    Dashed lines and markers indicate normalized error thresholds of $0.2$, $0.1$, and $0.05$.
    }
    \label{fig:reconstruction_error}
\end{figure}

\begin{figure*}[t]
\centering
    \includegraphics[width=0.8\linewidth]
    {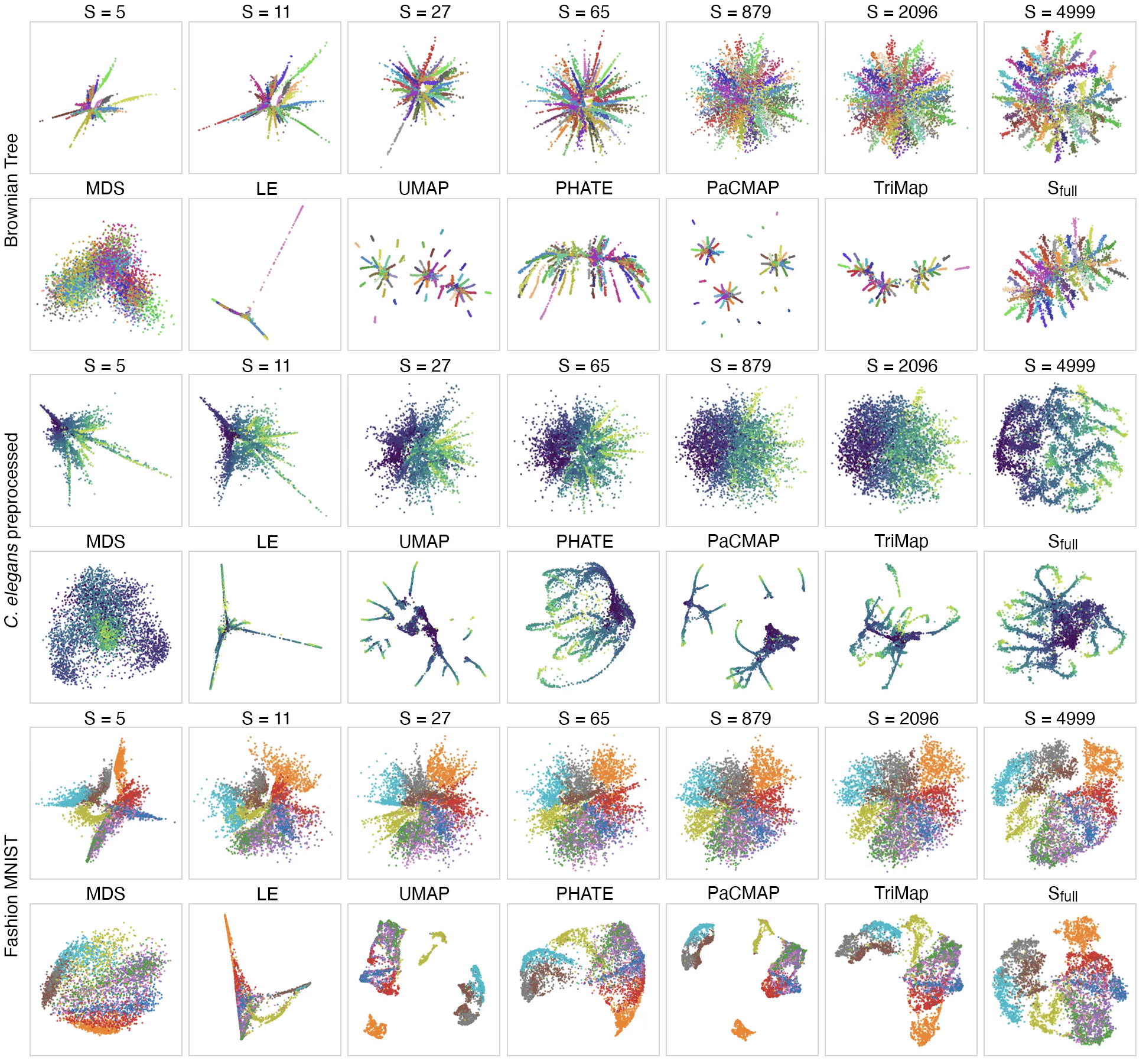}
    \caption{Qualitative comparison of embeddings across three datasets, colored by class (Brownian Tree, Fashion-MNIST) or embryo time (\textit{C.~elegans}).}
    \label{fig:progressive_embedding_results}
\end{figure*}

\vspace{3pt}\noindent\textbf{Observations.}
\cref{fig:reconstruction_error} shows the normalized reconstruction error for all seven datasets listed in \cref{tab:datasets}.
The curves reveal differences in spectral compressibility across the seven embeddings.
For most datasets, the normalized reconstruction error falls below $0.4$ within the first $30$ spectral modes.
Thus, these modes already reduce the discrepancy from the full-spectrum embedding by at least $60\%$ relative to the single-mode embedding.
After this initial decrease, most curves decay much more slowly, and closely approximating the full-spectrum embedding requires substantially more modes.
In the progressive embedding snapshots in \cref{fig:progressive_embedding_results}, discussed in \cref{sec:quantitative}, early stages capture prominent large-scale features, while later stages add finer geometric detail.

To compare spectral compressibility across datasets, we use normalized reconstruction-error thresholds of $0.2$, $0.1$, and $0.05$ and record the smallest $\nTopEigvals$ at which each threshold is reached. 
Across the seven datasets, the Swiss Roll embedding is the most spectrally compressible, reaching all three thresholds with the fewest modes.
This behavior can also be seen in the embeddings in \cref{fig:motivating_example}(a).
In contrast, the Brownian Tree and Multiscale Periodic Loop embeddings require much larger $\nTopEigvals$ to reach the thresholds.
For the Multiscale Periodic Loop, the error falls below $0.4$ with only two modes but decreases slowly thereafter.
The progressive embeddings already capture the overall circular structure (\cref{fig:motivating_example}(b)), while representing the fine oscillations requires modes near the high-frequency end of the spectrum.
The two single-cell embeddings also exhibit slow error decreases.
The unprocessed subset reaches $\error_\nTopEigvals \leq 0.2$ with fewer modes than the preprocessed dataset ($\nTopEigvals=3103$ vs. $\nTopEigvals=4440$), but both require a substantial fraction of the spectrum to reconstruct the final layout at this threshold.
For a chosen normalized reconstruction-error tolerance, analysts can use these curves to select the smallest number of low-frequency modes needed to reconstruct the final full-spectrum embedding.


\makeatletter
\def\thickhline{%
  \noalign{\ifnum0=`}\fi\hrule \@height \thickarrayrulewidth \futurelet
   \reserved@a\@xthickhline}
\def\@xthickhline{\ifx\reserved@a\thickhline
               \vskip\doublerulesep
               \vskip-\thickarrayrulewidth
             \fi
      \ifnum0=`{\fi}}
\makeatother
\setlength{\thickarrayrulewidth}{2\arrayrulewidth}

\begin{table}[t]
\vspace{5pt}
\centering
\caption{Datasets used in \cref{sec:results} and \cref{sec:cases}.}
\label{tab:datasets}
\small
\setlength{\tabcolsep}{3pt}
\renewcommand{\arraystretch}{0.9}
\vspace{-.7\abovecaptionskip}
\begin{tabular}{lrrl}
\thickhline
\textbf{Dataset} & $N$ & $M$ & \textbf{Used in} \\
\hline
Swiss Roll~\cite{tenenbaum2000global} & 5k & 3 & \cref{sec:reconstruction_error} \\
Multiscale Periodic Loop & 5k & 30 & \cref{sec:reconstruction_error,sec:usecase_mnist} \\
Brownian Tree~\cite{tenenbaum2000global,moon_PHATE_2019} & 5k & 50 & \cref{sec:reconstruction_error,sec:quantitative,sec:usecase_mnist} \\
MNIST~\cite{deng2012mnist} & 5k & 784 & \cref{sec:reconstruction_error,sec:complexity_analysis,sec:usecase_mnist}\\
Fashion-MNIST~\cite{xiao2017fashion} & 5k & 784 & \cref{sec:reconstruction_error,sec:quantitative,sec:usecase_mnist}\\
\textit{C.~elegans} neuron subset~\cite{packer2019lineage} & 5k & 20,220 & \cref{sec:reconstruction_error} \\
\textit{C.~elegans} (preprocessed)~\cite{packer2019lineage,monocle3trajectories} & 5k & 100 & \cref{sec:reconstruction_error,sec:quantitative,sec:usecase_singlecell,sec:usecase_mnist} \\
\thickhline
\end{tabular}
\end{table}

\begin{figure}[t!]
  \centering
  \includegraphics[width=1\columnwidth]{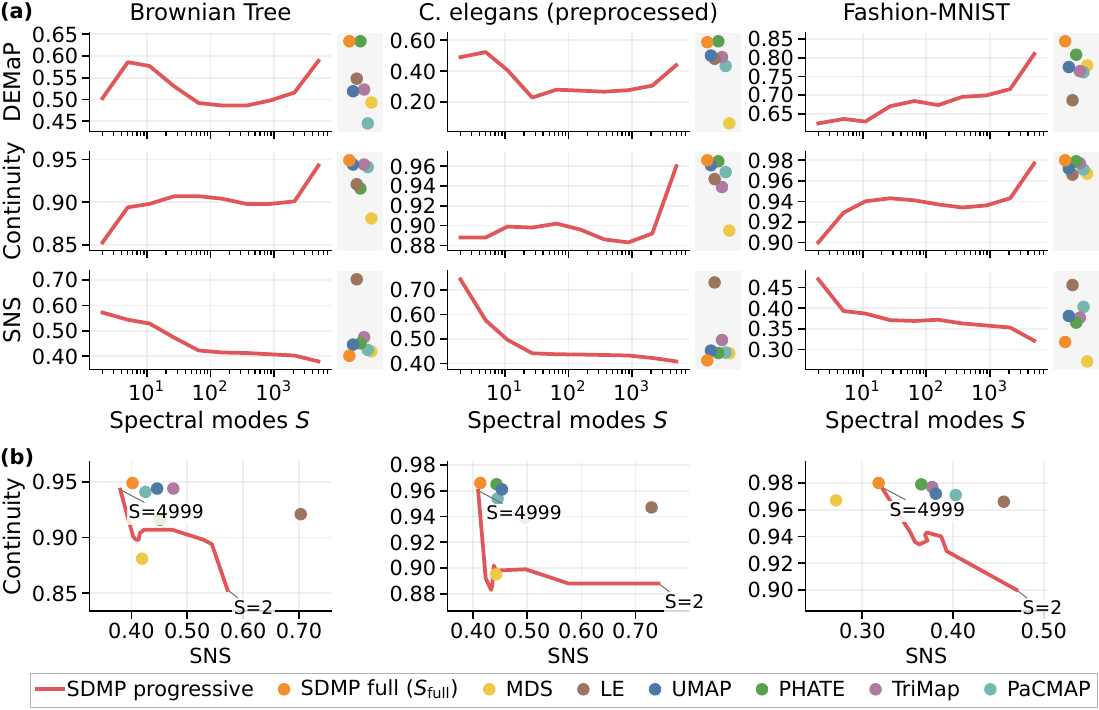}
  \caption{Quantitative evaluation across three datasets. (a) DEMaP, continuity, and SNS vs. spectral subspace size $S$, with baseline scores in shaded strips where horizontal jitter separates overlapping points. (b) Continuity vs. SNS (upper left is preferred). Lines connect progressive stages. SDMP full denotes non-progressive full-spectrum optimization.
  }
  \label{fig:quantitative_eval}
\end{figure}

\subsection{Embedding Quality Across Progressive Stages}
\label{sec:quantitative}

We evaluate how embedding quality changes across the progressive stages of \sysname{} and compare its embeddings with 
baselines, including MDS~\cite{torgerson1952multidimensional}, LE~\cite{belkin2003laplacian}, UMAP~\cite{mcinnes_umap_2020}, PHATE~\cite{moon_PHATE_2019}, TriMap~\cite{amid2019trimap}, and PaCMAP~\cite{wang2021pacmap}.
\cref{fig:progressive_embedding_results} provides visual comparisons of the resulting embeddings, while \cref{fig:quantitative_eval} shows the quantitative comparison.

Because quality measures emphasize different properties of nonlinear embeddings~\cite{jeon2023zadu, jung2025ghostumap2, machado2025necessary, jeon2025unveiling, wang2025dimension}, we use three complementary label-free measures to assess (1) geodesic-distance preservation, (2) local neighborhood preservation, and (3) global geometric distortion.
We intentionally avoid label-based or cluster-separation measures~\cite{jeon2023classes}, as such measures reward whether classes become visually compact in the embedding and can score high while substantially distorting manifold geometry.
(1) \textit{DEMaP}~\cite{moon_PHATE_2019}  computes the Spearman rank correlation between graph-geodesic distances in the original space and Euclidean distances in the embedding and favors manifold preservation.
(2) \emph{Scale-normalized stress} (SNS)~\cite{smelser2026scale} measures global geometric distortion while accounting for differences in scale.
(3) \emph{Continuity}~\cite{venna2006local} measures whether neighbors in the original space remain nearby after projection.
Higher DEMaP and continuity indicate better preservation, whereas lower SNS indicates less geometric distortion.
Sec.~B of the supplementary material provides the numerical scores plotted in \cref{fig:quantitative_eval}, along with additional measures of distance and neighborhood preservation: \emph{Spearman's $\rho$}~\cite{sidney1957spearman}, \emph{non-metric stress}~\cite{kruskal1964multidimensional}, and \emph{mean relative rank error}~\cite{lee2009rank}.
All measures except DEMaP are computed using the ZADU library~\cite{jeon2023zadu}.

\vspace{1pt}\noindent\textbf{Observations.}
At the final progressive stage, \sysname{} achieves lower SNS than all six baselines on Brownian Tree and \textit{C.~elegans}.
On Fashion-MNIST, only MDS achieves lower SNS.
Across all three datasets, continuity at the final progressive stage is within $0.005$ of the highest baseline score (\cref{fig:quantitative_eval}(a)).
The non-progressive full-spectrum run ($\stage{\mathrm{full}}$) achieves the highest continuity on all three datasets and the highest DEMaP on Fashion-MNIST.

\cref{fig:quantitative_eval}(b) summarizes local neighborhood preservation (continuity) and global geometric preservation (SNS), with the upper-left region indicating better performance on both measures.
Compared with $\stage{\mathrm{full}}$, the final progressive embeddings have lower SNS but slightly lower continuity on Brownian Tree and \textit{C.~elegans}.
Because both strategies ultimately use the same full spectral basis, these differences illustrate how the optimization path affects the local--global balance.
Across all three datasets, the final expansion from $\nTopEigvals=2096$ to $\nTopEigvals=4999$ improves both measures, producing the largest continuity gain between consecutive reported stages and further reducing SNS.
For Brownian Tree and \textit{C.~elegans}, the final expansion also reveals more clearly defined branches in \cref{fig:progressive_embedding_results}, coinciding with the largest continuity gains.

The intermediate stages reveal differences among the preservation measures.
For Brownian Tree and \textit{C.~elegans}, SNS decreases throughout the reported progression, while DEMaP rises from $S=2$ to $5$, falls at intermediate stages, and recovers toward the final stage. 
On \textit{C.~elegans}, the final progressive embedding also has lower SNS and lower DEMaP than $\stage{\mathrm{full}}$, highlighting the distinction between geometric distortion and geodesic-distance preservation.
Intermediate stages thus expose changes in structure preservation that evaluating only the final embedding would obscure.

\subsection{Scalability and Runtime Analysis}
\label{sec:complexity_analysis} 
We compare the runtime of the non-approximate implementation of \sysname{} with UMAP on a laptop equipped with an Apple M1 Pro chip and 16GB of unified memory. 
For $N=10{,}000$, $k=15$, and 200 optimization epochs, the explicit implementation takes approximately $1.2\times$ UMAP's runtime at $S=2$ and $7\times$ at $S=100$.
Sec.~C of the supplementary material provides the detailed comparison and computational complexity analysis.
The non-approximate implementation stores $O(NS)$ basis entries, requiring $O(N^2)$ storage at full spectrum.

The approximate strategy described in \cref{sec:progressive_strategy} avoids storing the full $O(N^2)$ spectral basis by caching only a user-specified number of spectral modes.
This enables progressive embedding of substantially larger datasets on the same 16\,GB laptop.
The approximate implementation reached the full-spectrum stage in 20 progressive steps on the full-size $70{,}000$ MNIST images in approximately 6 minutes and in 50 steps on the $1{,}306{,}127$ single-cell dataset~\cite{zheng2017massively} in approximately 4 hours. 
Both runs used 40 optimization epochs per step.

\section{Use Cases}
\label{sec:cases}

\begin{figure}[t]
\centering
    \includegraphics[width=1\linewidth]{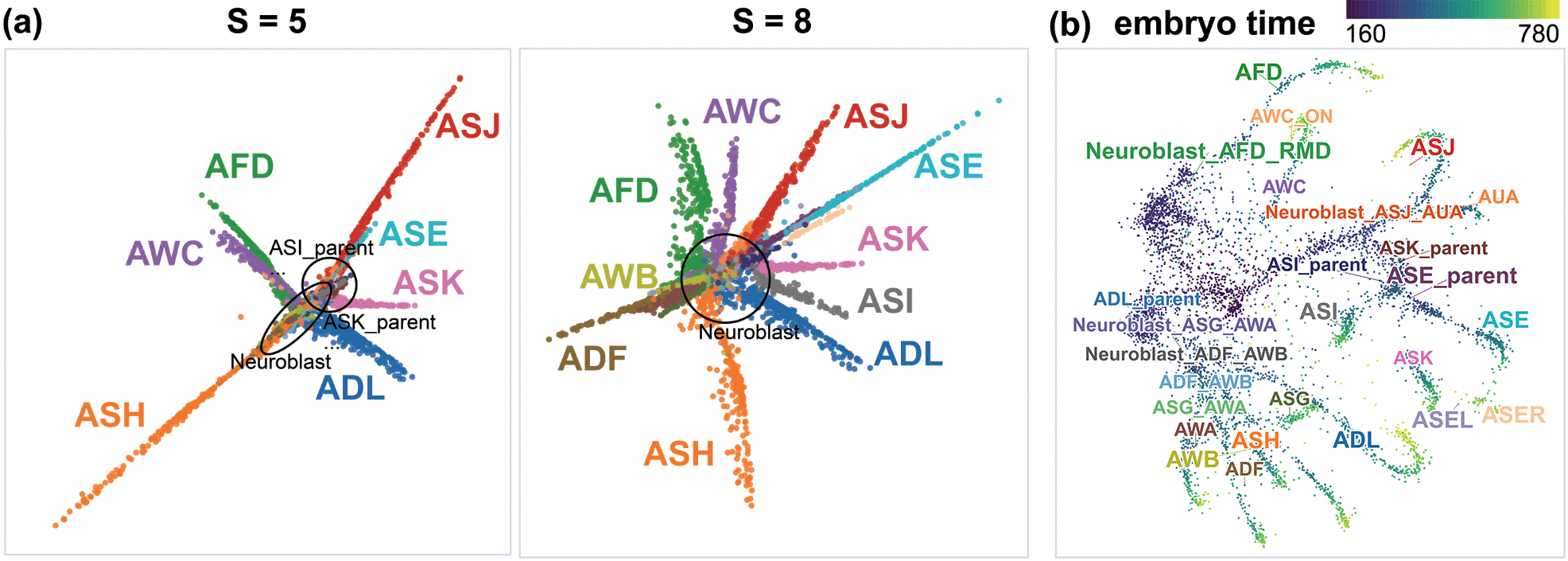} 
    \caption{Progressive embeddings of the preprocessed \textit{C.~elegans} data.
    (a) Early stages ($\nTopEigvals=5$ and $8$), colored by annotated cell type.
    (b) The final progressive stage ($\nTopEigvals= 4{,}999$), colored by estimated embryo time and labeled by cell type or lineage state.
    }
    \label{fig:single_cell}
\end{figure}

We present two use cases to illustrate how \sysname{} supports analysis in different settings.
The first case (\cref{sec:usecase_singlecell}) examines how progressive embeddings reveal biologically meaningful cell populations and developmental trajectories in the \textit{C.~elegans} embryogenesis data.
The second case (\cref{sec:usecase_mnist}) uses MNIST, Fashion-MNIST, and \textit{C.\,elegans} datasets to show how global and local explanations can provide cluster-level and instance-level insights.

\subsection{Progressive Embedding of \textit{C.~elegans} Embryogenesis}
\label{sec:usecase_singlecell}
DR methods are widely used in single-cell analysis~\cite{becht2019dimensionality,li2018mass, kobak2021initialization, monocle3trajectories, packer2019lineage}. 
Beyond the identification of clusters and labeling corresponding cell types~\cite{Hollt2016Cytosplore,vanUnen2017HSNE}, analyzing cell development as trajectories along the data manifold has gained attention~\cite{li2018mass, packer2019lineage,monocle3trajectories, moon_PHATE_2019, huguet2023heat}.
We analyze the single-cell RNAseq \textit{C.~elegans} embryogenesis dataset derived from Packer et al.~\cite{packer2019lineage}. 
This dataset is a strong benchmark, as it is well-studied and well-annotated, and data points lie on a dense manifold with small changes in between~\cite{fung2020cell, sivaramakrishnan2023transcript}.
Using the atlas's cell-type and lineage annotations, we examine when sensory neuron populations become visually distinct and how progenitor cells are positioned relative to their annotated descendants.
We use 20 logarithmically spaced stages ($\nTopEigvals=1,2,3,5,8, \ldots, 4999$) to inspect the early changes at closely spaced mode counts.
Our framework reveals two main findings.

\definecolor{AFD}{RGB}{76, 146, 78}
\definecolor{ASH}{RGB}{226, 127, 66}
\definecolor{ADL}{RGB}{56, 105, 163}
\definecolor{AWC}{RGB}{130, 98, 158}
\definecolor{ASJ}{RGB}{190, 64, 53}
\definecolor{ASE}{RGB}{87, 175, 194}
\definecolor{ASI}{RGB}{117, 117, 117}
\definecolor{AWB}{RGB}{177, 180, 83}
\definecolor{ASEParent}{RGB}{94, 55, 94}
\definecolor{ASEL}{RGB}{140, 138, 177}
\definecolor{ASER}{RGB}{234, 195, 152}
\definecolor{AFD}{RGB}{92, 151, 79}
\definecolor{NeuroblastAFDRMD}{RGB}{92, 149, 83}

First, several sensory neuron populations are visually distinct even when the embedding is restricted to low-frequency variation on the cell-neighborhood graph (\cref{fig:single_cell}(a)).
At $\nTopEigvals=5$, \textcolor{ASH}{\small\textsf{\textbf{ASH}}}, \textcolor{ADL}{\small\textsf{\textbf{ADL}}}, and \textcolor{ASJ}{\small\textsf{\textbf{ASJ}}} form distinct rays, while \textcolor{ASI}{\small\textsf{\textbf{ASI}}}, \textcolor{ASE}{\small\textsf{\textbf{ASE}}}, and \textcolor{AWB}{\small\textsf{\textbf{AWB}}} are easier to distinguish at $\nTopEigvals=8$.
Comparing these stages reveals which distinctions in the final embedding are already visible with five modes and which become clearer as the subspace expands.

Second, the full-spectrum embedding places annotated progenitor and descendant states near one another along individual branches (\cref{fig:single_cell}(b)).
For example, \textcolor{ASEParent}{\small \textsf{\textbf{ASE\_parent}}} lies near \textcolor{ASE}{\small\textsf{\textbf{ASE}}}, \textcolor{ASEL}{\small \textsf{\textbf{ASEL}}}, \textcolor{ASER}{\small \textsf{\textbf{ASER}}}; {\small \textcolor{NeuroblastAFDRMD}{\small \textsf{\textbf{Neuroblast\_AFD\_RMD}}}} lies near \textcolor{AFD}{\small\textsf{\textbf{AFD}}}.
Within each branch, analysts can use the embryo-time coloring to examine whether annotated progenitors occupy earlier estimated times than their descendants.

\begin{figure}[t]
\centering
    \includegraphics[width=1\linewidth]{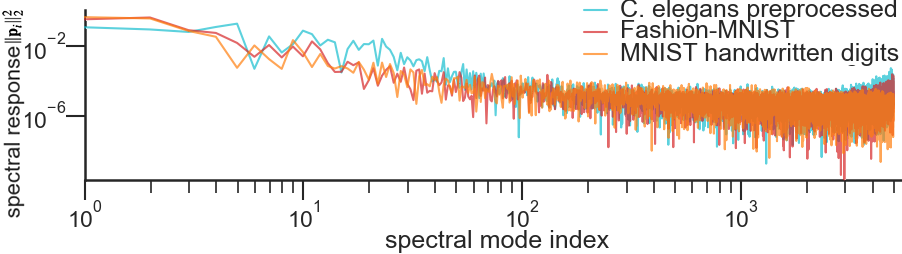} 
    \caption{
    Spectral response profile of the final-stage embeddings for real-world datasets in~\cref{tab:datasets}.
    }
    \label{fig:response_z}
\end{figure}

\subsection{\hspace{-4pt}Global and Local Spectral Explanations Across Datasets}
\label{sec:usecase_mnist}

We use the global and local visual explanations from \cref{sec:visual_explaination} to examine which visible patterns in an embedding are supported by smoothly varying structure on the data graph and which depend on finer-scale variation.
We compare these contributions across datasets and examine how they vary among classes and individual points.

\subsubsection{Global Explanation}
The spectral response profile in \cref{fig:response_z} complements the reconstruction error.
The reconstruction error summarizes how many modes are needed to approximate the full-spectrum embedding (cf.~\cref{sec:reconstruction_error}), whereas the response profile reveals which individual modes contribute most strongly to the layout.

\vspace{3pt}
\noindent \textbf{Key insight 1: Visible embedding patterns can be supported by different graph scales.}
We compare MNIST handwritten digits, Fashion-MNIST, and \textit{C.~elegans} in \cref{fig:response_z}.
For each mode $i$, the response profile plots the coefficient magnitude $\|\mathbf{\ProjMatElem}_i\|_2^2$, normalized to sum to one over all modes, where $\mathbf{\ProjMatElem}_i$ is the corresponding row of the projection matrix $\ProjMat{}_{\nTopEigvals}$ in \cref{eq:core_equation}.
The magnitude is squared because of Parseval's identity mentioned in~\cref{sec:layout_optimization}, such that each value is the fraction of the layout carried by that mode.
A larger magnitude indicates that the mode contributes more strongly to the embedding.
MNIST and Fashion-MNIST concentrate their responses in the smoothest graph modes. 
Their first ten modes account for approximately 95\% and 92\% of the total coefficient magnitude, respectively.
In contrast, the first ten modes account for only approximately 69\% for \textit{C.~elegans}, indicating that its embedding draws on a broader range of graph patterns.
The broader response in \textit{C.~elegans} motivates a closer look at which cell populations account for the contributions of later modes.

\begin{figure}[t]
\centering
\includegraphics[width=1\linewidth]{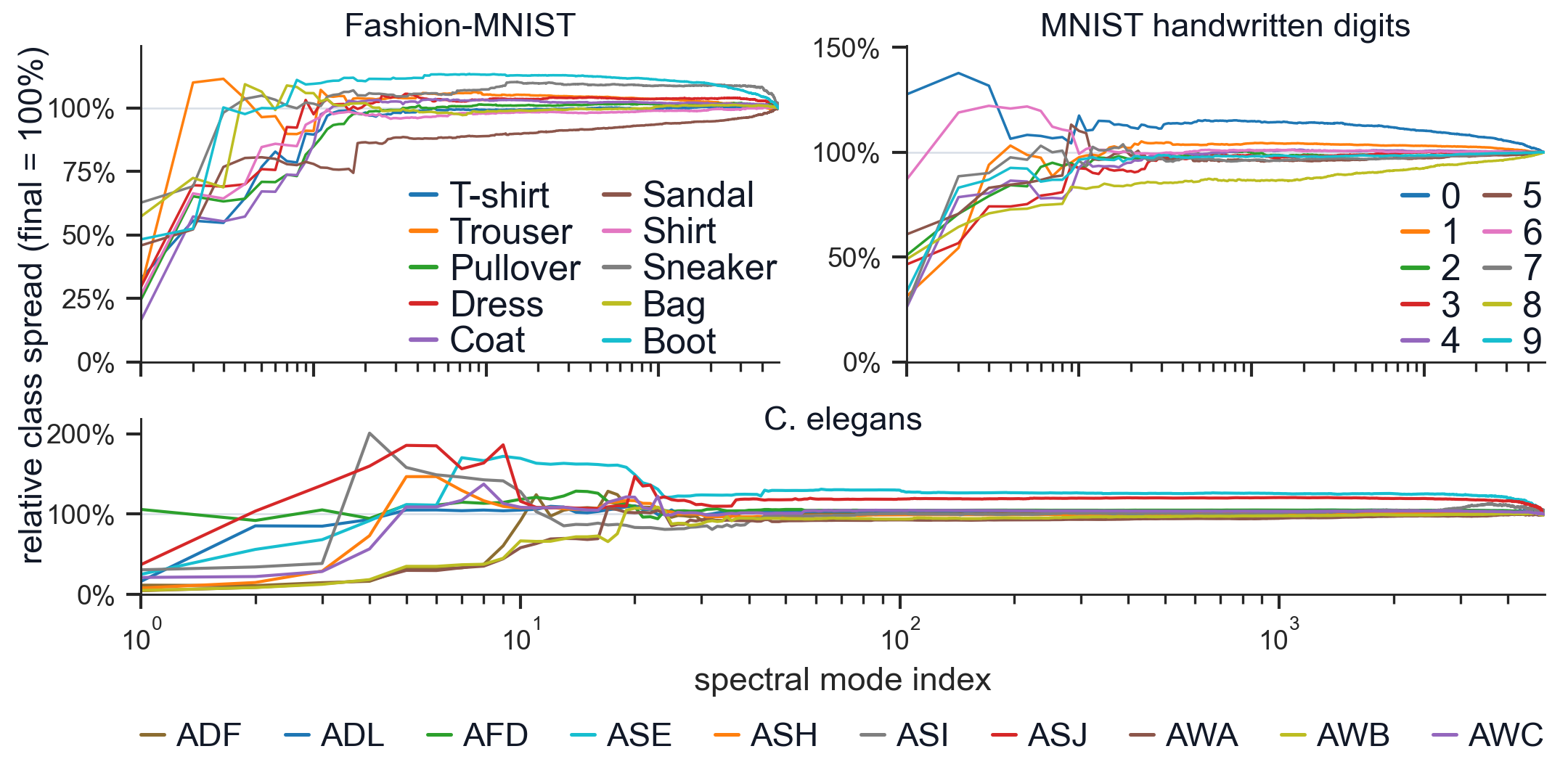}
    \caption{
    Within-class spread as spectral modes are added for Fashion-MNIST (top-left), MNIST (top-right), and \textit{C.~elegans} with ten selected cell types (bottom).
    The $x$-axis shows the spectral subspace size $\nTopEigvals$, and each curve shows one class's spread relative to its spread in the full-spectrum embedding.
    A value of $100\%$ indicates the same spread as the full-spectrum embedding, while values above and below $100\%$ indicate greater dispersion and greater compactness, respectively.
    }
    \label{fig:expl_class}
\end{figure}

\subsubsection{Local Explanation}
\label{sec:case_local_explanation}

Complementing the progressive embeddings in \cref{fig:progressive_embedding_results}, \cref{fig:expl_class} shows how within-class compactness changes as spectral modes are added.
For each subspace size $\nTopEigvals$, we measure within-class spread as the root-mean-square distance of the class's points from their centroid.
We normalize each class's spread by its value in the full-spectrum embedding. 
Thus, $100\%$ indicates the same spread as in the full-spectrum embedding, while values above and below $100\%$ indicate greater dispersion and greater compactness, respectively.

\begin{figure}[t]
\centering
    \includegraphics[width=1\linewidth]{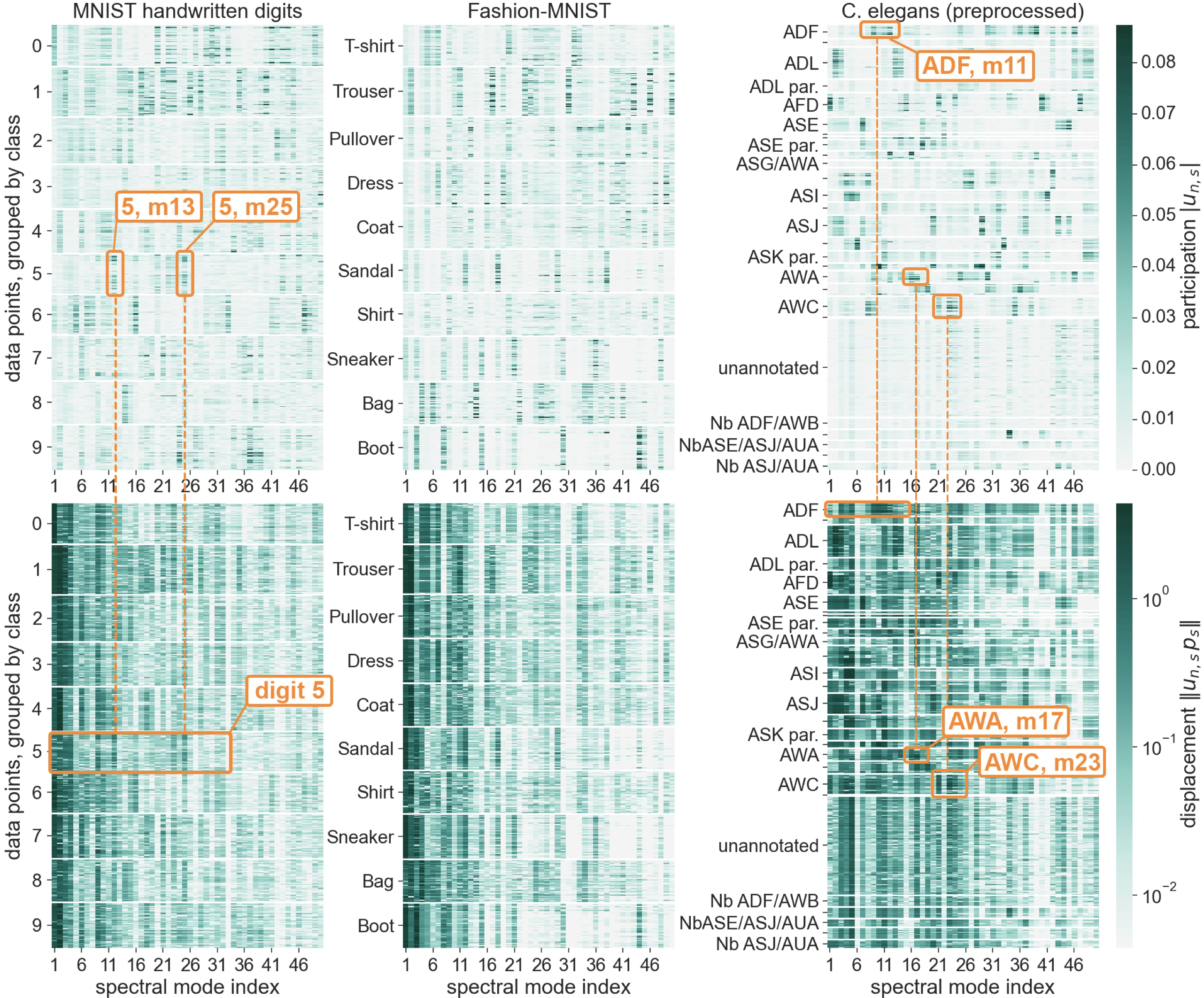}
    \caption{
    Point-level participation magnitudes $|\EigvecElement{n,i}|$ (top row) and displacement magnitudes $\norm{\EigvecElement{n,i}\mathbf{\ProjMatElem}_i}_2$ (bottom row) over the first 50 nontrivial spectral modes.
    Within each heatmap, columns represent spectral modes, and rows represent points.
    Rows are grouped by class label and cell annotation to expose group-level patterns.
    }
    \label{fig:expl_point}
\end{figure}

\cref{fig:expl_point} shows each point's participation and displacement magnitudes in the first 50 nontrivial spectral modes for three different datasets. 
Visualizations covering all spectral modes and the other datasets are provided in the supplementary material.
For point $n$ and mode $i$, the participation and displacement magnitudes are $|\EigvecElement{n,i}|$ and $\norm{\EigvecElement{n,i}\mathbf{\ProjMatElem}_i}_2$, respectively.
The displacement magnitude essentially equals the participation magnitude scaled by the learned coefficient magnitude $\norm{\mathbf{\ProjMatElem}_i}_2$.
Thus, comparing the paired heatmaps reveals how optimization weights graph-derived patterns when forming the embedding.

\vspace{1pt}
\noindent 
\textbf{Key insight 2: Globally dominant graph scales may not characterize individual groups.}
Comparing the global response profile in \cref{fig:response_z} with the within-class spread curves in \cref{fig:expl_class} reveals a difference between dataset-level and class-level behavior.
For MNIST, the first ten modes carry approximately $95\%$ of the total response (\cref{fig:response_z}), indicating that these smooth modes dominate the overall embedding.
However, as seen in \cref{fig:expl_class}, with ten modes, digits $0$ and $8$ are at $118\%$ and $83\%$ of their final spread, respectively, and remain at $115\%$ and $86\%$ after $50$ modes.
As a result, subsequent modes do not affect all classes uniformly; they tighten one cluster while loosening the other.
After $50$ modes, eight MNIST classes have spreads between $90\%$ and $110\%$ of their final values, whereas only slightly more than half of the \textit{C.~elegans} classes fall within this range.
Thus, when preserving subtle class-level compactness or dispersion is important, analysts should not select a reduced spectral subspace from the global response alone, but should also examine how individual classes change as modes are added.

\vspace{1pt}
\noindent 
\textbf{Key insight 3: Higher spectral modes can make concentrated contributions to particular groups.}
At the dataset level, the displacement magnitudes in \cref{fig:expl_point} generally decrease as the mode index increases, consistent with the global response profiles in \cref{fig:response_z}.
However, this trend is not uniform across groups or points.
For example, for MNIST digit $5$, the strong participation around modes $13$ and $25$ produces correspondingly concentrated displacement magnitudes for the same points.
For \textit{C.~elegans}, the mean displacement magnitudes of {\small \textsf{\textbf{ADF}}}, {\small \textsf{\textbf{AWA}}}, and {\small \textsf{\textbf{AWC}}} show peaks at modes 11, 17, and 23, respectively.
Thus, the broad response tail observed for \textit{C.~elegans} does not represent a uniform refinement of the embedding. 
Instead, different modes contribute primarily to different cell populations.
These local explanations help identify groups or points affected by individual modes and determine whether higher modes remain important for a population of interest.

These results collectively show that a global spectral summary alone may be insufficient for interpreting an embedding or selecting a spectral subspace. 
By making mode contributions explicit, \sysname{} enables analysts to identify, at the dataset, group, and point levels, the graph scales and regions supporting patterns relevant to their analytical task. 
Starting from an apparent split in the final embedding, analysts can trace points on either side across earlier stages to see whether the groups are already distinct when only low-frequency modes are available.
They can then compare the original feature values of points on opposite sides to investigate what distinguishes the groups.

\section{Discussion and Future Work}
\label{sec:discussion}
The goal of this paper is not to propose yet another DR algorithm that simply produces ``better'' final embeddings. 
Rather, we present a different formulation of graph-based DR that is inspired by graph signal processing~\cite{shuman2013emerging, defferrard2016spectralgnn, balestriero2022contrastive} and run optimization~\cite{mcinnes_umap_2020, damrich_umap_2021} under a linearly combined spectral coordinate system, keeping the graph frequency structure explicit throughout the embedding process. 
The embeddings from \sysname{} become structured objects that support multiscale construction and intrinsic explanation of their formation.

Several directions could further extend this formulation. 
First, instead of selecting a fixed truncation of the spectrum, one could parameterize the spectral response by a learnable filter~\cite{defferrard2016spectralgnn, wang2022powerfulgnn} applied to $\ProjMat$. 
This would provide a more expressive alternative to hard subspace selection while preserving the interpretability of the spectral parameterization.
Another direction is related to the known issue of the reliability of $k$NN graphs in neighbor embeddings~\cite{maple, saidi2025recovering} and the corresponding graph augmentation techniques.
Our proposed framework may offer a useful perspective for revealing and explaining the effect of graph augmentation in graph-based DR~\cite{moon_PHATE_2019, huguet2023heat}. 
Under \sysname{}, different objectives and graph modifications may amplify or suppress different parts of the spectrum, thereby favoring different balances between coarse global organization and local separation.
Studying these methods in a common spectral language may provide a principled way to better understand graph-based DR methods. 
Overall, we see this work as a step toward a more structured view of nonlinear dimensionality reduction, where the embedding is not only a final scatterplot but also a decomposable and analyzable representation of graph structure.

\section{Conclusion}
\label{sec:conclusion}
We introduced \sysname{}, a spectral framework that combines Laplacian graph frequencies with neighborhood-preserving cross-entropy optimization, thereby optimizing embeddings in controlled spectral subspaces.
By progressively expanding these spectral subspaces, SDMP produces coarse-to-fine snapshots that make the global--local trade-off explicit and inspectable.
Across the quantitative evaluations and use cases on synthetic, image, and single-cell data, \sysname{} preserves manifold continuity and provides global and local explanations. 
More broadly, this work reframes DR embeddings as multiscale, decomposable graph-signal objects.

\section*{Acknowledgments}
The authors wish to thank the anonymous reviewers for their valuable feedback.
This work was partially supported by the Gender and Diversity Group of ELLIIT, a strategic research environment in Sweden.

\raggedbottom 

\bibliographystyle{abbrv-doi-hyperref}
\bibliography{00b-ref}

\end{document}